\documentclass[letterpaper, 10 pt, conference]{ieeeconf}  % Comment this line out if you need a4paper

\IEEEoverridecommandlockouts                              % This command is only needed if 
\usepackage{cite}
\usepackage{epsfig} % for postscript graphics files
\usepackage{mathptmx} % assumes new font selection scheme installed
\usepackage{times} % assumes new font selection scheme installed
\usepackage{amsmath,amssymb,amsfonts}
\usepackage{algorithmic}
\usepackage{graphicx} 
\usepackage{textcomp}
\usepackage{xcolor}
\usepackage{subcaption}
\usepackage{glossaries} % use \glsresetall after abstract
\usepackage{hyperref}
\def\BibTeX{{\rm B\kern-.05em{\sc i\kern-.025em b}\kern-.08em
    T\kern-.1667em\lower.7ex\hbox{E}\kern-.125emX}}

\begin{document}

% \title{\LARGE \bf
% STAG: A Sparse Traversability-Aware Graph Representation from Grid-Based Costmaps for Robotic Navigation
% }

% \author{Gabriel Garcia$^{1}$, Stephanie Aravecchia$^{2}$ and Miguel Olivares Mendez% <-this % stops a space
% \thanks{*This work was not supported by any organization}% <-this % stops a space
% \thanks{$^{1}$Albert Author is with Faculty of Electrical Engineering, Mathematics and Computer Science,
%         University of Twente, 7500 AE Enschede, The Netherlands
%         {\tt\small albert.author@papercept.net}}%
% \thanks{$^{2}$Bernard D. Researcheris with the Department of Electrical Engineering, Wright State University,
%         Dayton, OH 45435, USA
%         {\tt\small b.d.researcher@ieee.org}}%
% }

\newif\ifanonymized
\anonymizedfalse      % <-- set to \ anonymizedfalse for review submission

\ifanonymized
\title{STAG: A Sparse Traversability-Aware Graph Representation from Grid-Based Costmaps for Robotic Navigation
\thanks{This research was funded in whole, or in part, by Anonymous research institution, grant reference [Anonymous]. For the purpose of open access, and in fulfilment of the obligations arising from the grant agreement, the author has applied a Creative Commons Attribution 4.0 International (CC BY 4.0) license to any Author Accepted Manuscript version arising from this submission.}
}
\author{Anonymous 1$^{1}$, Anonymous 2$^{2}$, Anonymous 3$^{1}$%
\thanks{$^{1}$ From Anonymous center, Anonymous}%
\thanks{$^{2}$ From Anonymous center, Anonymous}%
}
\hypersetup{
    hidelinks,
    pdfauthor={}
}

\else
\title{STAG: A Sparse Traversability-Aware Graph Representation from Grid-Based Costmaps for Robotic Navigation
\thanks{This research was funded in whole, or in part, by the Luxembourg National Research Fund (FNR), grant reference [CAESAR-XR/Gabriel/17679211]. For the purpose of open access, and in fulfilment of the obligations arising from the grant agreement, the author has applied a Creative Commons Attribution 4.0 International (CC BY 4.0) license to any Author Accepted Manuscript version arising from this submission.}
}
\author{Gabriel Manuel Garcia$^{1}$, St{\'e}phanie Aravecchia$^{2}$, Miguel Angel Olivares-Mendez$^{1}$%
\thanks{$^{1}$ From University of Luxembourg, Luxembourg}%
\thanks{$^{2}$ From IRL Georgia Tech-CNRS, Metz, France}%
}
\fi

\maketitle
\thispagestyle{empty}
\pagestyle{empty}

%%%%%%%%%%%%%%%%%%%%%%%%%%%%%%%%%%%%%%%%%%%%%%%%%%%%%%%%%%%%%%%%%%%%%%%%%%%%%%%%
\begin{abstract}
Autonomous rovers navigating large unstructured environments need efficient global planning that accounts for terrain traversability. However, searching dense grid-based costmaps becomes computationally expensive as the mapped area grows. We introduce STAG, a Sparse Traversability-Aware Graph that converts costmaps into compact graphs. STAG combines a medial-axis topological backbone, representative nodes for homogeneous traversability regions, and transition nodes near strong traversability gradients. Edges encode geometry and traversability to account for path length and terrain difficulty. We compare A* on STAG and dense grids using synthetic cave maps, mine maps and the DARPA CERBERUS dataset. Across five benchmark categories comprising 203 map instances and 101,200 queries, STAG reduces median planning time by \(3.4\times\) to \(9.9\times\) and peak query memory by \(2.1\times\) to \(15.4\times\), with median relative path-length differences of \(-2.9\%\) and \(+7.6\%\). STAG offers a compact representation for global planning, trading dense-grid traversability optimality for faster, less memory-intensive search.
\end{abstract}

% \begin{keywords}
% Motion and Path Planning, Mining Robotics, Robotics in Under-Resourced Settings
% \end{keywords}

%%%%%%%%%%%%%%%%%%%%%%%%%%%%%%%%%%%%%%%%%%%%%%%%%%%%%%%%%%%%%%%%%%%%%%%%%%%%%%%%
\section{INTRODUCTION}

Autonomous robots operating in large unstructured environments require internal terrain representations that support safe and efficient path planning. This requirement is particularly important for terrestrial and planetary caves, lava tubes, mines, and subterranean environments, where robots must navigate over long distances with limited onboard computation and limited communication with human operators. In such environments, the representation must capture not only which areas are free or occupied, but also how difficult the terrain is to traverse.

Grid-based traversability costmaps are widely used for this purpose because they provide a simple and spatially detailed representation of obstacles and terrain difficulty. However, planning directly on a dense grid becomes increasingly expensive as the explored area grows. Each traversable cell becomes a potential planning state, and repeated path queries over large maps can lead to high computation times and memory requirements \cite{kuffner2004grid,c24}. This creates a practical limitation for long-range autonomous missions, where global planning must be performed repeatedly on resource-constrained platforms.

In this paper, we address the problem of reducing the computational cost of path planning over grid-based traversability costmaps while preserving the terrain information needed for risk-aware global navigation. Instead of planning directly over all traversable cells, we convert the costmap into a sparse graph that preserves the main free-space connectivity and selected traversability structure.

Existing graph-based representations provide useful tools for reducing planning complexity. Topological and skeleton-based methods, such as generalized Voronoi graphs and distance-field skeletons, compactly represent the connectivity of free space~\cite{c1}, \cite{c3}. Such abstractions have been used successfully in large-scale exploration systems, including subterranean navigation frameworks~\cite{c6}, \cite{c11}. However, purely topological graphs mainly encode geometry and do not explicitly preserve terrain-cost variations. Traversability-aware graph methods reduce the search space while encoding terrain risk, but their representations are often coupled to particular sensing modalities, connection strategies, or dedicated planners~\cite{c7,c20,c22}.

To address these limitations, we introduce STAG, a Sparse Traversability-Aware Graph representation. STAG converts a grid-based traversability costmap into a compact multi-source graph composed of three complementary elements: a medial-axis topological backbone, region nodes that summarize homogeneous traversability areas, and transition nodes placed near strong traversability gradients. The resulting graph stores geometric and terrain-related attributes on its edges, allowing standard graph-search algorithms to evaluate both path length and terrain difficulty without searching over the full grid.

The contribution of this paper is a sparse traversability-aware representation for global path planning over costmaps. More specifically, we make the following claims:

\begin{itemize}
    \item STAG reduces the number of planning states compared with the original grid-based traversability costmap.
    \item Using the same A* search formulation and matched start-goal queries, STAG reduces query time and memory relative to dense-grid search.
    \item STAG preserves global path geometry closely, while exposing a measurable trade-off between sparse search efficiency and dense-grid traversability optimality.
\end{itemize}

These claims are evaluated on 200 generated cave and mine maps, one long synthetic mine, and two DARPA-derived traversability maps by comparing grid-based A* planning with STAG-based A* planning. The evaluation focuses on representation size, planning time, peak query memory, path length, and accumulated traversability cost. The results show that STAG provides a lightweight global planning representation for repeated path queries over traversability maps, while highlighting the expected loss of fine terrain-cost optimality caused by sparse abstraction.

\section{Related Work}

Environment representations for robotic navigation must balance spatial detail, terrain awareness, and planning efficiency. Dense costmaps preserve local obstacle and traversability information, whereas graph-based representations reduce the number of states considered during repeated planning. STAG addresses the latter problem by converting an existing two-dimensional traversability costmap into a sparse graph while retaining selected geometric and terrain-cost structure.

\subsection{Grid-Based and Multi-Resolution Representations}

Grid costmaps represent free space, obstacles, and navigation costs on a regular lattice and are commonly searched using Dijkstra's algorithm or A*~\cite{hart1968astar,lu2014layered}. Their principal advantage is that the planner can directly exploit the complete spatial and terrain-cost resolution of the map. However, the number of search states increases with map extent and resolution, which can make repeated global queries expensive~\cite{kuffner2004grid}. Multi-resolution methods reduce this burden by coarsening portions of the search space while retaining finer resolution where necessary~\cite{c24}. Such approaches nevertheless remain coupled to a grid hierarchy and its refinement policy. STAG instead constructs an explicit reusable graph whose vertices are selected from the geometry and traversability structure of the input costmap.

\subsection{Skeleton and Topological Graphs}

Topological representations reduce planning complexity by preserving the principal connectivity of free space with fewer states. Generalized Voronoi graphs provide obstacle-clear routes through an environment~\cite{c1}, while skeleton-based approaches derive sparse roadmaps from distance fields or medial-axis structures~\cite{c2,c3,c9,c10}. Incremental Euclidean signed-distance-field systems such as Voxblox provide geometric and clearance information that can support these abstractions~\cite{c4}. Sparse topological graphs have also been used for large-scale and subterranean exploration, including multi-robot systems operating under limited communication~\cite{c5,c6,c11,c12}.

These methods demonstrate that free-space topology can substantially reduce representation size while retaining long-range connectivity. Their graph construction is nevertheless driven primarily by geometry, clearance, or exploration structure. Areas with similar geometry but different traversal difficulty may therefore remain indistinguishable. STAG adopts a medial-axis backbone for geometric connectivity, but augments it with nodes selected directly from the traversability field and stores terrain-related statistics on graph edges.

\subsection{Traversability-Aware and Region-Based Graphs}

Traversability-aware planners associate terrain risk or cost with graph nodes and edges. TRG-Planner constructs a traversal-risk graph and jointly uses it with a risk-aware planning strategy for navigation in unstructured terrain~\cite{c7}. Other approaches construct topological graphs through generalized-Voronoi sampling~\cite{c8}, derive weighted traversability graphs from three-dimensional terrain data~\cite{c20}, or associate traversability with connections between visually defined places for long-range off-road navigation~\cite{c22}. Hierarchical terrain-aware systems further combine multiple graph or map resolutions to support exploration and navigation decisions~\cite{c15}. These works demonstrate the value of incorporating terrain information into sparse planning structures, but their representations are commonly coupled to a particular sensing modality, sampling procedure, or specialized planner.

Region-based abstractions provide another means of reducing representation size. Topological maps can represent convex or spatially coherent free-space regions as graph elements~\cite{c13,c19}, while graph-based segmentation methods estimate traversable terrain from LiDAR or other sensor data~\cite{c17,c18}. The latter primarily address perception and terrain estimation rather than compression of an already available planning costmap, and are therefore complementary to STAG.

STAG specifically targets an existing two-dimensional traversability costmap. It combines three deterministic node sources: a medial-axis backbone, representatives of connected homogeneous traversability regions, and nodes near strong traversability transitions. The resulting graph retains explicit edge geometry, clearance, and traversability attributes and can be searched using a standard graph-search algorithm. This separates the contribution of the environment representation from that of a dedicated perception or planning pipeline.

% ===============================================================================

\section{Methodology}

\begin{figure}[t]
    \centering
    \includegraphics[width=0.8\columnwidth]{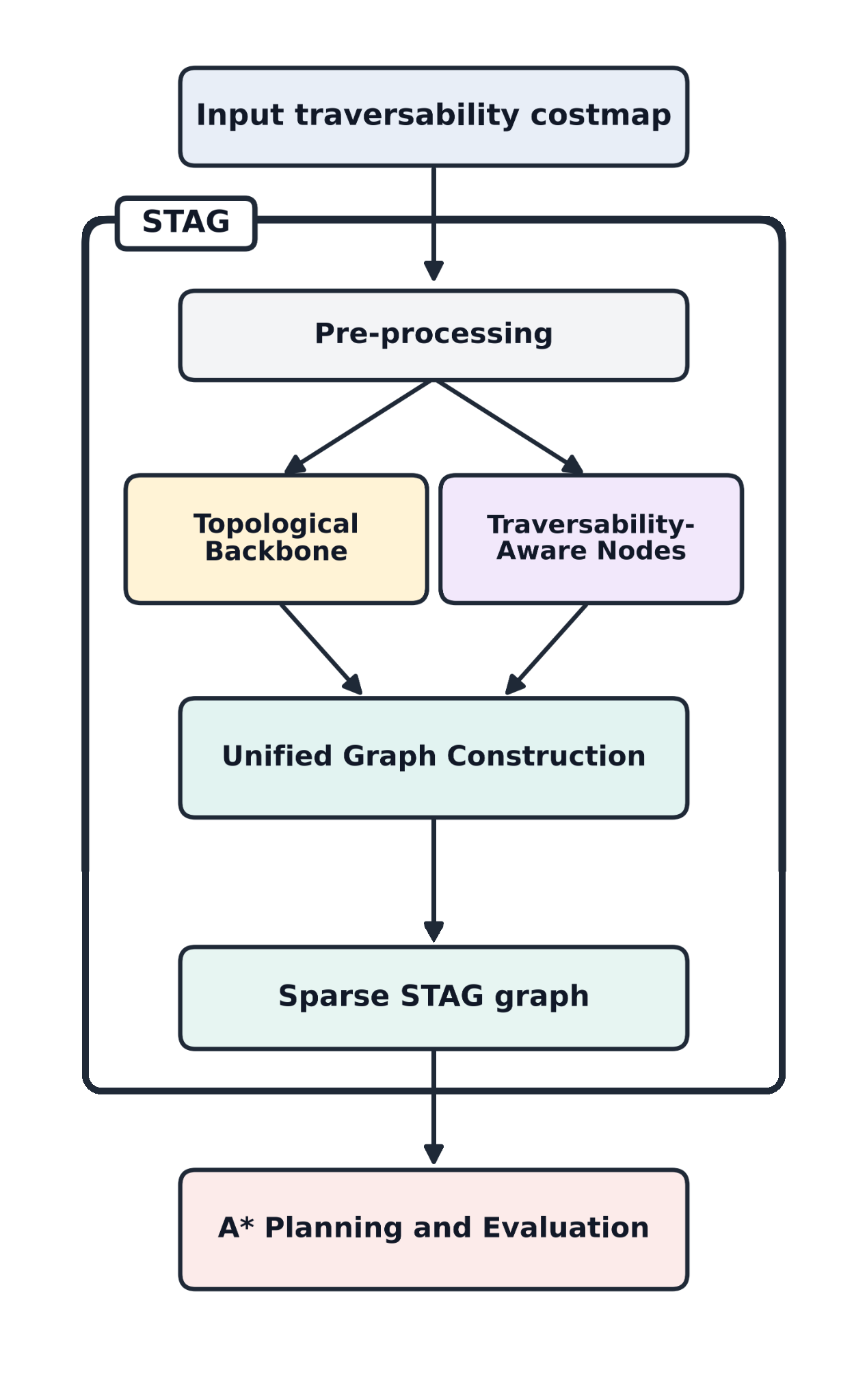}
    \caption{Overview of the STAG pipeline. An input traversability costmap is preprocessed to extract a topological backbone and traversability-aware region and transition nodes, which are combined into a unified sparse graph for A* planning and evaluation.}
    \label{fig:overview}
\end{figure}

Figure~\ref{fig:overview} summarizes STAG pipeline. Given a grid-based traversability costmap, STAG constructs a sparse graph \(G=(V,E)\) whose vertices come from three sources: a topological backbone, homogeneous traversability regions, and high-gradient terrain transitions. Edges store geometric and terrain-related attributes, allowing graph-search algorithms to evaluate both distance and terrain difficulty without expanding all traversable grid cells.

\subsection{Input Representation}

The input costmap is first converted into a normalized traversability score field \(S:\Omega\rightarrow[0,1]\), defined over the set of grid coordinates \(\Omega\subset\mathbb{Z}^{2}\). Higher values of \(S(x,y)\) indicate easier terrain, whereas lower values indicate more difficult terrain. Obstacle and unknown cells are marked as invalid during this preprocessing step. A binary traversability mask \(M:\Omega\rightarrow\{0,1\}\) is then defined as:

\begin{equation}
M(x,y)=
\begin{cases}
1, & S(x,y)\geq\tau_S \text{ and the cell is valid},\\
0, & \text{otherwise},
\end{cases}
\label{eq:traversability_mask}
\end{equation}

where \(\tau_S\) is the minimum traversability score accepted for planning. As defined in~\eqref{eq:traversability_mask}, cells below this threshold, together with obstacle and unknown cells, are treated as non-traversable. The corresponding terrain penalty is defined as follows, such that higher values of \(P(x,y)\) represent more difficult terrain:

\begin{equation}
P(x,y)=1-S(x,y),
\label{eq:terrain_penalty}
\end{equation}

\subsection{Topological Backbone}

The first component of STAG captures the global structure of the environment independently of local terrain cost. Instead of representing all traversable cells, the free space is reduced to a topological skeleton.

An Euclidean distance transform is computed over the traversable mask \(M\). This distance field gives, for each traversable cell, its distance to the closest obstacle and is used to extract the medial axis of the free space. The medial axis provides a compact skeleton of the environment while preserving its connectivity.

The skeleton is then converted into a graph. Skeleton endpoints and junctions become graph vertices, while the chains connecting them form
graph edges. Each edge stores its geometric length and clearance statistics derived from the distance field. As shown in Fig.~\ref{fig:topological_graph}, the resulting backbone preserves the main free-space connectivity while substantially reducing the number of planning states.

% \begin{figure}[t]
%     \centering
%     \includegraphics[width=\columnwidth]{fig/topology_graph.png}
%     \caption{Topological backbone extracted from the binary traversability mask. Gray and white cells denote non-traversable and traversable space, respectively. Thin dark lines represent medial-axis branches; yellow markers indicate endpoints, dark-red markers indicate junctions, and hollow gray markers denote intermediate corridor cells used to describe the edge geometry.}
%     \label{fig:topological_graph}
% \end{figure}

The endpoints, junctions, and additional nodes introduced to represent closed skeleton branches form the topological node set \(\mathcal{N}_{\mathrm{top}}\).

\subsection{Traversability-Aware Nodes}

The topological graph preserves connectivity but does not fully describe how terrain difficulty varies across the environment. STAG therefore adds two types of traversability-aware nodes: region nodes and transition nodes.

\subsubsection{Region Nodes}

Region nodes summarize areas with similar traversability values. The traversability field \(S\) is discretized into a finite set of traversability classes. Connected components are then extracted independently within each class.

Let \(R_k\) denote the \(k\)-th connected region:

\begin{equation}
R_k =
\left\{
\mathbf{p}_{k,1},
\mathbf{p}_{k,2},
\ldots,
\mathbf{p}_{k,N_k}
\right\},
\label{eq:region_cells}
\end{equation}

As defined in~\eqref{eq:region_cells}, \(\mathbf{p}_{k,i}=[x_{k,i},y_{k,i}]^\mathsf{T}\) is the two-dimensional coordinate of the \(i\)-th cell in \(R_k\), and \(N_k=|R_k|\) is the number of cells in that region.

The centroid of \(R_k\) is computed as

\begin{equation}
\mathbf{c}_k =
\frac{1}{N_k}
\sum_{i=1}^{N_k}
\mathbf{p}_{k,i}.
\label{eq:region_centroid}
\end{equation}

Because the centroid in~\eqref{eq:region_centroid} is not necessarily located on a valid grid cell, the representative node is selected as the cell in \(R_k\) that is closest to it:

\begin{equation}
\mathbf{p}^{\star}_k =
\underset{\mathbf{p}\in R_k}{\operatorname{arg\,min}}
\left\|
\mathbf{p}-\mathbf{c}_k
\right\|_2^2.
\label{eq:region_representative}
\end{equation}

Let \(K\) denote the number of connected traversability regions. The
complete set of region nodes is

\begin{equation}
\mathcal{N}_{\mathrm{reg}}
=
\left\{
\mathbf{p}^{\star}_k
\right\}_{k=1}^{K}.
\label{eq:region_node_set}
\end{equation}

Equation~\eqref{eq:region_node_set} retains one valid representative cell for each connected component of a traversability class, selected by~\eqref{eq:region_representative} as the cell nearest to its centroid.
The resulting node distribution is shown in Fig.~\ref{fig:region_nodes}.

% \begin{figure}[t]
%     \centering
%     \includegraphics[width=\columnwidth]{fig/map_region_cleanup_medoid_nodes.png}
%     \caption{Region nodes extracted from homogeneous traversability areas. Each connected region is represented by the traversable cell closest to its centroid.}
%     \label{fig:region_nodes}
% \end{figure}

\subsubsection{Transition Nodes}

Region nodes summarize homogeneous terrain but do not explicitly represent locations where traversability changes sharply. STAG therefore
introduces transition nodes near strong local variations of the traversability field.

Let \(\widetilde{S}\) denote the smoothed traversability field. Its gradient magnitude is:
\begin{equation}
G(x,y)
=
\left\|\nabla \widetilde{S}(x,y)\right\|_2
=
\sqrt{
\left(\frac{\partial \widetilde{S}}{\partial x}\right)^2+
\left(\frac{\partial \widetilde{S}}{\partial y}\right)^2
},
\label{eq:gradient_magnitude}
\end{equation}

where large values of \(G(x,y)\) indicate strong local changes in terrain difficulty. Transition-node candidates are selected among the
traversable local maxima of~\eqref{eq:gradient_magnitude} that exceed a gradient threshold. Candidates are processed in decreasing gradient
magnitude, while a minimum spatial separation is enforced to avoid redundant nodes.

The resulting set of transition nodes is denoted by:
\begin{equation}
\mathcal{N}_{\mathrm{tr}}
=
\left\{
\mathbf{p}_{\mathrm{tr},j}
\right\}_{j=1}^{n_{\mathrm{tr}}},
\label{eq:transition_node_set}
\end{equation}

where \(n_{\mathrm{tr}}\) is the number of retained transition nodes. Their distribution is illustrated in Fig.~\ref{fig:transition_nodes}.

% \begin{figure}[t]
%     \centering
%     \includegraphics[width=\columnwidth]{fig/map_output_G_nodes.png}
%     \caption{Transition nodes placed on high-gradient areas of the traversability map. These nodes represent local changes in terrain difficulty.}
%     \label{fig:transition_nodes}
% \end{figure}

% Together, region nodes and transition nodes provide a compact semantic description of the terrain: region nodes encode homogeneous areas, while transition nodes encode changes between them.

\subsection{Unified Graph Construction}

Finally, STAG combines the topological node set \(\mathcal{N}_{\mathrm{top}}\), the region node set defined in~\eqref{eq:region_node_set}, and the transition node set defined in~\eqref{eq:transition_node_set} into a unified graph \(G=(V,E)\).

\begin{equation}
V =
\mathcal{N}_{\mathrm{top}}
\cup
\mathcal{N}_{\mathrm{reg}}
\cup
\mathcal{N}_{\mathrm{tr}},
\label{eq:unified_node_set}
\end{equation}

where \(\mathcal{N}_{\mathrm{top}}\) also includes any topological nodes
introduced when splitting existing edges.

% Each traversability-aware node is connected to nearby valid elements of the topological graph. Candidate connectors are accepted only if they remain inside traversable space and satisfy the required clearance constraints. When a semantic node connects to the interior of a topological edge, the edge is split at the connector location so that the final graph remains consistent.

% If a traversability aware node connects to the middle of an edge joining nodes u and v, STAG inserts a new node s and replaces (u,v)
Each region or transition node in \(\mathcal{N}_{\mathrm{reg}}\cup\mathcal{N}_{\mathrm{tr}}\) is connected to nearby topological nodes or to projected points on topological edges. A connector is retained only if its path remains within traversable space and satisfies the minimum-clearance requirement. When a connector ends inside a topological edge, a new topological node is inserted at that location, and the original edge is replaced by the two resulting edge segments. This ensures that every connector terminates at a graph node while preserving the connectivity of the topological backbone.

For each edge \(e\in E\), STAG stores its geometric length, minimum and mean clearance, and minimum and mean traversability scores. These attributes allow graph-search algorithms to account for both distance and terrain difficulty.

\begin{figure*}[!t]
    \centering
    \captionsetup[sub]{font=footnotesize,justification=centering}

    % First row
    \begin{minipage}[t][7.7cm][t]{0.49\textwidth}
        \vspace{0pt}
        \centering
        \includegraphics[
            height=6.3cm,
            trim=11mm 8mm 0 4mm,
            clip
        ]{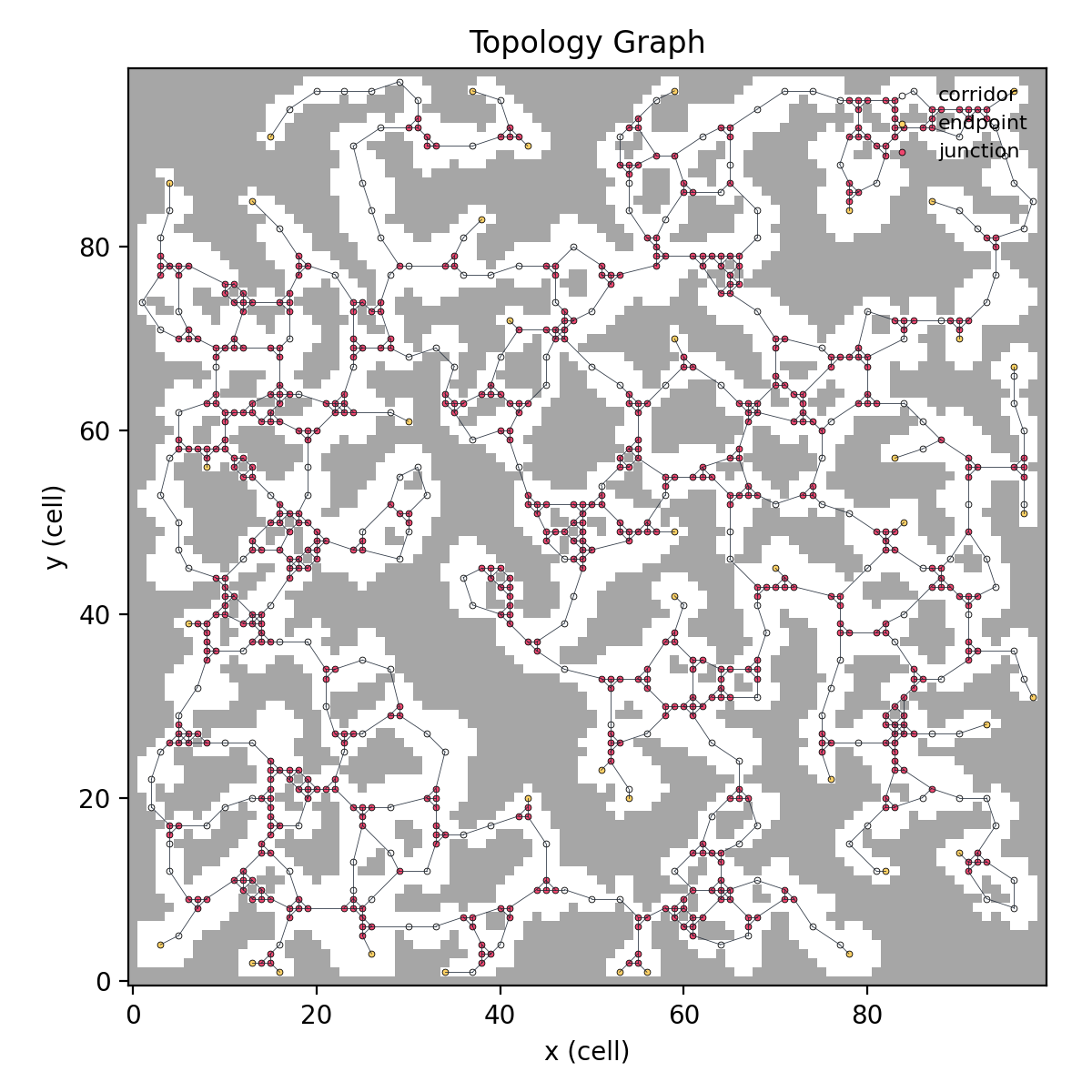}
        \subcaption{Topological backbone. Yellow and dark-red markers
        denote endpoints and junctions, respectively.}
        \label{fig:topological_graph}
    \end{minipage}
    \hfill
    \begin{minipage}[t][7.7cm][t]{0.49\textwidth}
        \vspace{0pt}
        \centering
        \includegraphics[
            height=6.3cm,
            trim=0 20mm 0 10mm,
            clip
        ]{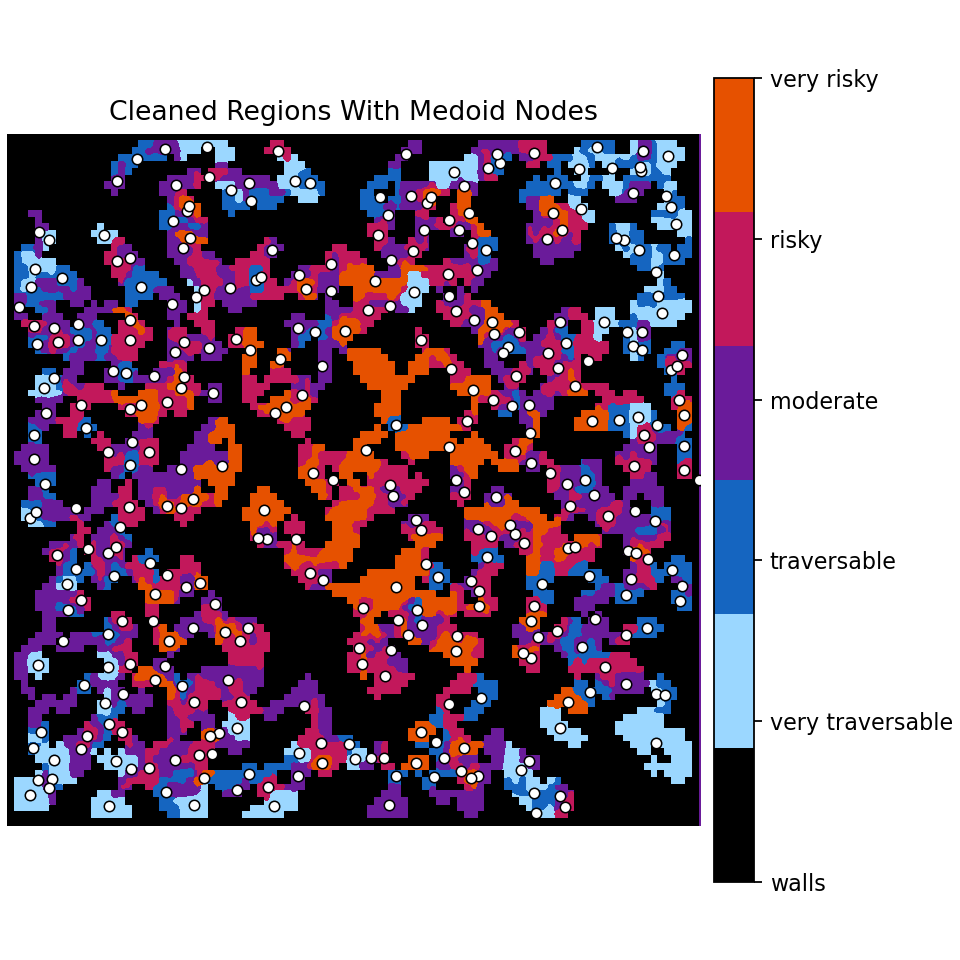}
        \subcaption{Region representatives for connected areas belonging to the same traversability class.}
        \label{fig:region_nodes}
    \end{minipage}

    \vspace{0.4em}

    % Second row
    \begin{minipage}[t][7.7cm][t]{0.49\textwidth}
        \vspace{0pt}
        \centering
        \includegraphics[
            height=6.3cm,
            trim=0mm 0 0 0,
            clip
        ]{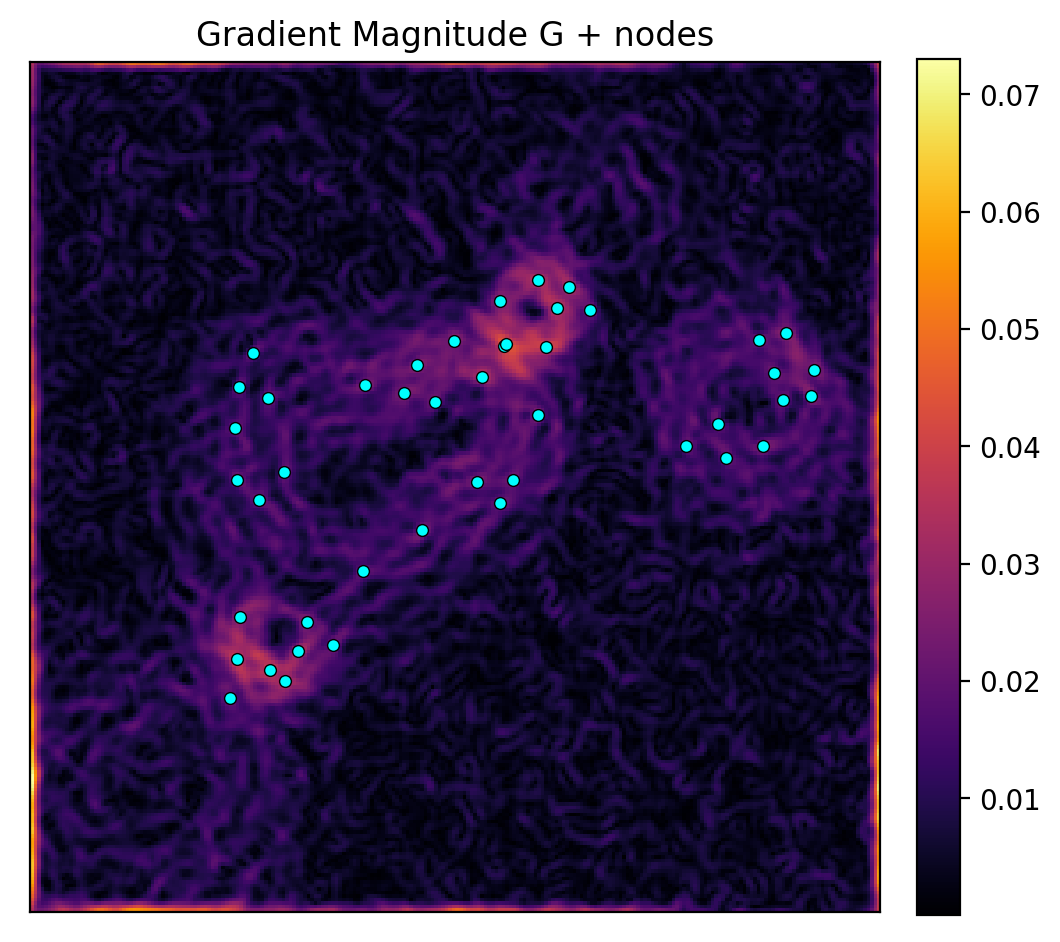}
        \subcaption{Illustrative gradient-magnitude field on a separate costmap, with transition nodes retained near strong local maxima.}
        \label{fig:transition_nodes}
    \end{minipage}
    \hfill
    \begin{minipage}[t][7.7cm][t]{0.49\textwidth}
        \vspace{0pt}
        \centering
        \includegraphics[
            height=6.3cm,
            trim=11mm 8mm 5mm 4mm,
            clip
        ]{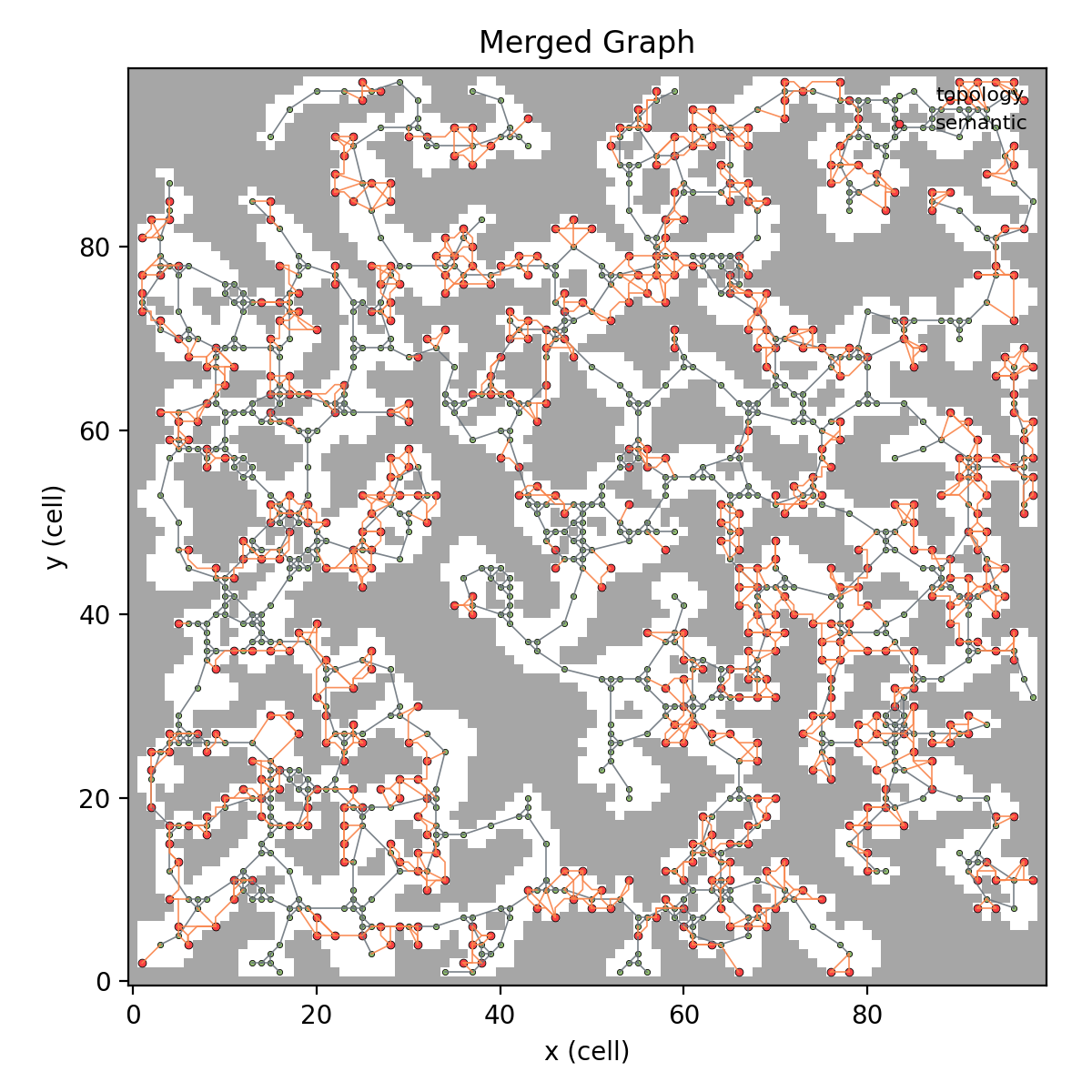}
        \subcaption{Unified STAG graph combining the topological,
        region, and transition node sets.}
        \label{fig:unified_graph}
    \end{minipage}

    \caption{Representative stages of STAG construction. Panels (a), (b), and (d) show the topological backbone, traversability-region representatives, and unified STAG graph, respectively, for the same traversability costmap. Panel (c) uses a separate costmap to illustrate the gradient-magnitude field and the selection of transition nodes near strong local traversability changes.}
    \label{fig:stag_construction}
\end{figure*}

The unified representation defined in~\eqref{eq:unified_node_set} is shown in Fig.~\ref{fig:unified_graph}.

% \begin{figure}[t]
%     \centering
%     \includegraphics[width=\columnwidth]{fig/graph.png}
%     \caption{Unified STAG graph overlaid on the traversability mask. Gray and white cells denote non-traversable and traversable space, respectively. Orange markers denote topological nodes, green markers denote traversability-aware nodes, and dark lines denote graph edges.}
%     \label{fig:unified_graph}
% \end{figure}

The resulting graph preserves the main connectivity of the environment while retaining the traversability information required for path planning. 
% Compared with the original dense grid, STAG reduces the number of planning elements and enables path search directly on a sparse graph representation.

% The STAG algorithm proceeds as follows:
% \begin{algorithmic}[1]
% \STATE Extract traversable mask \(M\) and traversability score \(S\) from costmap \(C\)
% \STATE Compute distance transform and medial-axis skeleton of \(M\)
% \STATE Convert skeleton junctions, endpoints, and branches into a topological graph
% \STATE Extract region nodes from connected homogeneous traversability regions
% \STATE Extract transition nodes from local maxima of \(\|\nabla S\|\)
% \STATE Connect traversability-aware nodes to valid nearby graph elements
% \STATE Store edge length, clearance, and traversability attributes
% \end{algorithmic}

% \begin{figure*}[t]
%     \centering
%     \includegraphics[
%         width=\textwidth,
%         keepaspectratio
%     ]{fig/stag_construction_summary_equal_panels.png}
%     \caption{Construction of the unified STAG graph. 
%     (a) Topological backbone. 
%     (b) Region nodes classified according to traversability. 
%     (c) Transition nodes. 
%     (d) Unified STAG graph combining topological and semantic information.}
%     \label{fig:stag_construction}
% \end{figure*}

\section{Experiments}

The proposed method is not a new path planning algorithm, but a new environment representation for path planning. The experiments therefore evaluate whether planning on STAG is more efficient than planning on the original grid-based traversability costmap, while maintaining an acceptable path-quality trade-off.

The benchmark compares two representations: the original grid costmap and the proposed STAG graph. In both cases, A* is used for the same tuple [start, goal] queries. This isolates the effect of the representation from the effect of the planner. All results are reported as medians over the accepted start-goal queries for each map. For each environment, the input costmap is loaded once and the corresponding STAG graph is constructed once. Both representations are then reused for all planning queries. Timing and memory measurements are taken only during the planning query, including start-goal attachment for STAG, and do not include graph construction. This separates the offline representation-building cost from the repeated online planning cost.
% No comparison with dedicated traversability graph planners is included in this evaluation.

\subsection{Benchmark Setup}

% The evaluation is performed on synthetic mine-like environments and mine-derived traversability maps. Synthetic environments provide repeatable conditions with different map sizes and terrain distributions. Mine-derived maps provide more realistic cases with irregular tunnel structures and traversability variations representative of subterranean exploration.

% For each map, valid start-goal pairs are sampled from traversable cells. A pair is accepted only if both endpoints are traversable and a valid grid path exists. Each accepted query is solved once on the grid representation and once on the STAG representation. The same sampled start-goal pairs are used for all comparisons on a given map.

% The STAG graph is generated once per map before running the planning queries. For each query, the start and goal positions are attached to the graph before search. The reported planning time includes this attachment step, since it is required when planning from arbitrary positions. Although the implementation supports optional KD-tree acceleration for graph attachment, all reported experiments use the lightweight deterministic attachment procedure.

The evaluation comprises 203 traversability-map instances grouped into five benchmark categories. The Synthetic caves and Synthetic mines categories each contain 100 generated environments. In addition, the benchmark includes one Long synthetic mine environment and two DARPA-derived environments, represented at resolutions of \(0.5\,\mathrm{m}\) and \(0.25\,\mathrm{m}\), respectively. The synthetic environments provide repeatable test cases with varying map sizes and terrain distributions, whereas the DARPA-derived maps contain irregular tunnel geometries and traversability variations representative of subterranean exploration. The two real-world costmaps were derived from sensor data recorded by Team CERBERUS during the DARPA Subterranean Challenge Final Event at the Louisville Mega Cavern~\cite{cerberus_dataset}.

For each of the 100 Synthetic caves and 100 Synthetic mines environments, 500 planning queries are evaluated, resulting in 50,000 queries for each category. The Long synthetic mine is evaluated using 200 queries, while each DARPA-derived environment is evaluated using 500 queries. The complete benchmark therefore contains 101,200 planning queries.

For each map, start-goal pairs were sampled until the target number of matched successful queries was reached. A pair was included only when both grid-based A* and STAG-based A* returned a valid path. The query counts reported in Table~\ref{tab:stag_results} therefore denote matched successful queries, and all grid and STAG metrics are computed over the same retained pairs.

For STAG planning, the start and goal cells generally do not coincide with existing graph nodes. Each endpoint is therefore attached to a node in the largest connected component of the graph before A* search. Euclidean distances to the candidate graph nodes are computed and the nodes are tested in increasing distance order until a traversable straight-line connection is found. This attachment procedure is applied to both \(s_i\) and \(g_i\), and its computation time is included in the reported STAG query time.

The representative synthetic mine was fixed by a random seed selected before benchmark generation.

Experiments ran on Ubuntu 24.04.4 LTS on an Intel Xeon W-10855M CPU \(2.80\,\mathrm{GHz}\)). Per-query CPU time was measured using \texttt{time.process\_time()}, and peak Python allocations using \texttt{tracemalloc} around each query. Measurements exclude graph, adjacency, and search-index construction, serialization, and persistent storage; STAG query measurements include start-goal attachment. Grid and STAG A* were executed independently on the same matched start-goal pair.

\subsection{Cost Function and Metrics}

For both representations, A* minimizes a cost combining geometric length and terrain difficulty. The cell-wise terrain penalty is defined in~\eqref{eq:terrain_penalty}. For an edge \(e\), let \(L(e)\) denote its geometric length and let

\begin{equation}
\overline{P}(e)
=
\frac{1}{L(e)}
\int_e P(\mathbf{x})\,\mathrm{d}s
\label{eq:mean_edge_penalty}
\end{equation}

denote the length-normalized mean terrain penalty sampled along the edge. The planning cost of \(e\) is then

\begin{equation}
J(e)
=
L(e)\left(1+\lambda\overline{P}(e)\right),
\label{eq:planning_cost}
\end{equation}

where \(\lambda\) controls the influence of terrain difficulty. The same value, \(\lambda=4\), is used for the grid and STAG representations.

For a path \(\mathcal{P}=(e_1,e_2,\ldots,e_m)\), the geometric length and accumulated traversability penalty are

\begin{equation}
L(\mathcal{P})
=
\sum_{i=1}^{m}L(e_i),
\qquad
C_P(\mathcal{P})
=
\sum_{i=1}^{m}L(e_i)\overline{P}(e_i).
\label{eq:path_metrics}
\end{equation}

The cumulative planning cost is therefore

\begin{equation}
J(\mathcal{P})
=
\sum_{i=1}^{m}J(e_i)
=
L(\mathcal{P})+\lambda C_P(\mathcal{P}).
\label{eq:path_cost}
\end{equation}

Relative path-length and accumulated traversability-penalty differences
are computed for each matched query as

\begin{equation}
\small
\Delta L =
\frac{L_{\mathrm{STAG}}-L_{\mathrm{grid}}}
     {L_{\mathrm{grid}}}
\times 100\%,
\quad
\Delta C_P =
\frac{C_{P,\mathrm{STAG}}-C_{P,\mathrm{grid}}}
     {C_{P,\mathrm{grid}}}
\times 100\%.
\label{eq:relative_differences}
\end{equation}

Positive values of \(\Delta L\) and \(\Delta C_P\) indicate that the
STAG path is longer or accumulates a larger terrain penalty than the
corresponding grid path, respectively.

The evaluation reports representation size, planning time, peak query memory, geometric path length, and accumulated traversability penalty. Representation size is measured by the number of traversable grid cells for the baseline and by the numbers of graph nodes and edges for STAG. Planning time and peak query memory are measured per query.

\section{Results and Discussion}

\begin{figure*}[t]
    \centering
    \includegraphics[width=\textwidth]{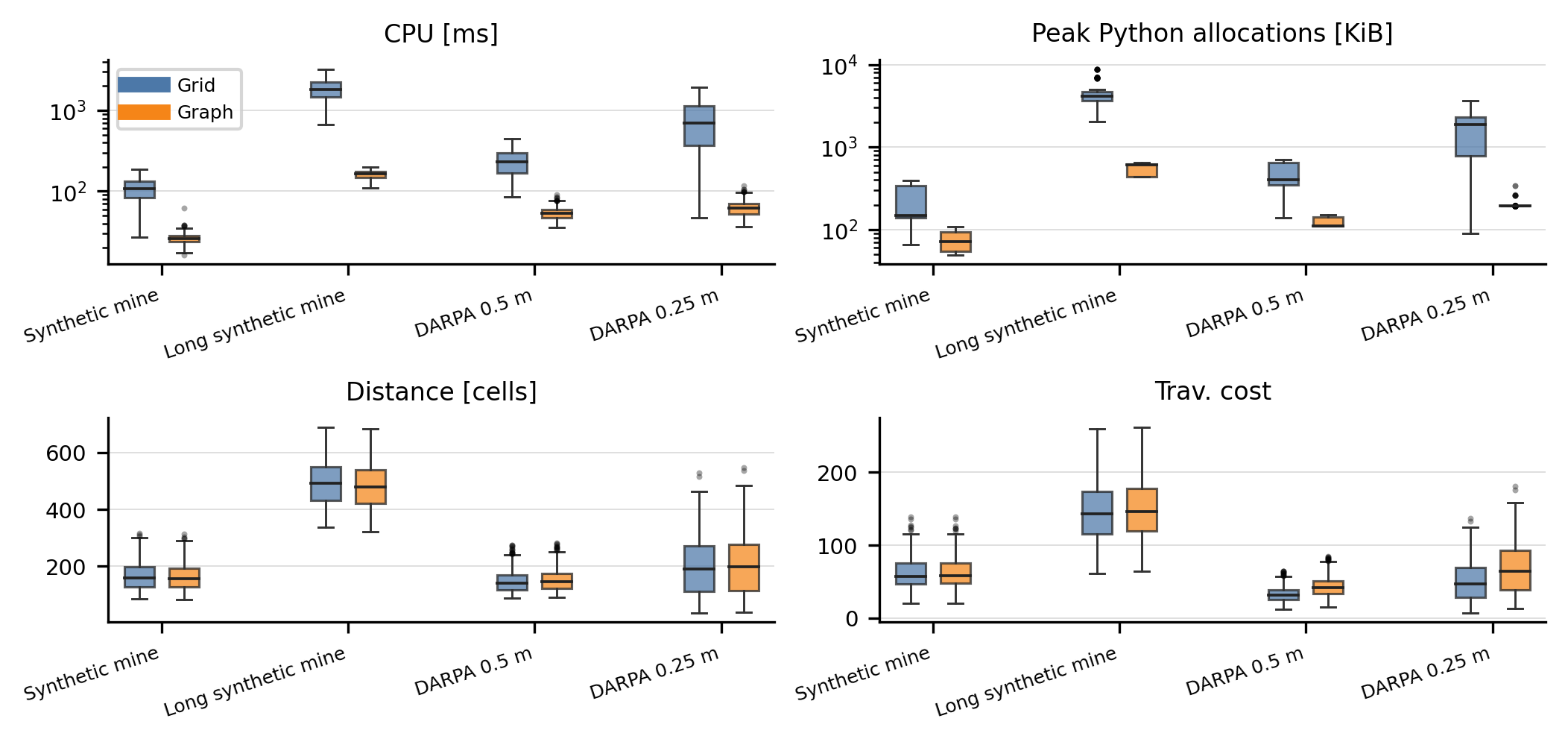}
    \caption{Query-level distributions of grid-based and STAG-based A* planning metrics for a representative synthetic mine, the Long synthetic mine, and the two DARPA-derived environments. The figure complements the aggregate five-category results in Table~\ref{tab:stag_results}. STAG reduces CPU time and peak query memory, while the path-quality trade-off varies with the environment.}
    \label{fig:boxplot_summary}
\end{figure*}

\begin{table*}[t]
\centering
\caption{Grid-based and STAG-based A* comparison. CPU time and peak
memory correspond to the planning query, including endpoint attachment
for STAG. CPU and memory entries report the median $\pm$ one standard
deviation. For Synthetic caves and Synthetic mines, the standard deviation
is computed over the 100 environment-level medians; for the three
single-map environments, it is computed over planning queries. The CPU
speed-up and memory-reduction factors are computed from the reported
medians as the grid value divided by the corresponding STAG value.
Positive values of $\Delta L$ and $\Delta C_P$ indicate that the STAG path
is longer or has a higher terrain penalty than the grid path.}
\label{tab:stag_results}

\resizebox{\textwidth}{!}{
\begin{tabular}{lrrrr|rrr|rrr|rr}
\hline
\textbf{Environment} &
\textbf{Open cells} &
\textbf{Queries} &
\textbf{Nodes} &
\textbf{Edges} &
\multicolumn{3}{c|}{\textbf{CPU time [ms]}} &
\multicolumn{3}{c|}{\textbf{Peak memory [KiB]}} &
$\boldsymbol{\Delta L}$ &
$\boldsymbol{\Delta C_P}$ \\
& & & & &
\textbf{Grid} &
\textbf{STAG} &
\textbf{Factor} &
\textbf{Grid} &
\textbf{STAG} &
\textbf{Factor} &
\textbf{[\%]} &
\textbf{[\%]} \\
\hline

Synth caves &
14,502 &
50,000 &
1,550 &
7,755 &
\(658.0 \pm 294.6\) &
\(\mathbf{82.5 \pm 31.8}\) &
8.0\(\times\) &
\(1679.2 \pm 605.4\) &
\(\mathbf{203.4 \pm 113.2}\) &
8.3\(\times\) &
+7.6 &
+6.9 \\

Synth mines &
5,918 &
50,000 &
1,416 &
7,536 &
\(264.8 \pm 48.9\) &
\(\mathbf{77.2 \pm 14.8}\) &
3.4\(\times\) &
\(407.7 \pm 76.7\) &
\(\mathbf{198.1 \pm 44.3}\) &
2.1\(\times\) &
+0.9 &
+3.4 \\

Long synth mine &
24,132 &
200 &
3,448 &
18,829 &
\(1797.2 \pm 500.6\) &
\(\mathbf{181.2 \pm 41.2}\) &
9.9\(\times\) &
\(4158.3 \pm 1493.4\) &
\(\mathbf{603.4 \pm 96.8}\) &
6.9\(\times\) &
-2.9 &
+2.0 \\

DARPA 0.5 m &
3,688 &
500 &
967 &
4,834 &
\(260.6 \pm 95.8\) &
\(\mathbf{68.9 \pm 37.4}\) &
3.8\(\times\) &
\(400.2 \pm 162.5\) &
\(\mathbf{105.8 \pm 17.1}\) &
3.8\(\times\) &
+6.0 &
+50.9 \\

DARPA 0.25 m &
14,756 &
500 &
1,274 &
5,897 &
\(744.8 \pm 478.8\) &
\(\mathbf{98.2 \pm 82.7}\) &
7.6\(\times\) &
\(1883.3 \pm 1000.4\) &
\(\mathbf{122.0 \pm 14.4}\) &
15.4\(\times\) &
+6.2 &
+38.2 \\

\hline
\end{tabular}
}
\end{table*}

\subsection{Representation Compression}

Figure~\ref{fig:boxplot_summary} visually compares grid- and STAG-based query metrics on four representative environments, while Table~\ref{tab:stag_results} reports the aggregate results over all five benchmark categories. STAG reduces the number of planning states in all evaluated maps. As seen in Table \ref{tab:stag_results}, in the synthetic cave environments, the graph contains 1550 nodes compared with 14502 traversable grid cells. In the long synthetic mine map, it contains 3,448 nodes compared with 24,132 traversable cells. The strongest reduction is observed in the DARPA 0.25 m map, where STAG uses 1,274 nodes for 14,756 traversable cells.

Although the number of graph edges can remain high, each edge represents a larger traversable segment rather than a single grid-cell transition. The resulting graph therefore provides a sparse planning structure while preserving the main connectivity of the environment. This supports the first claim of the paper: STAG reduces the number of planning states compared with the original grid-based traversability costmap.

\subsection{Planning Time and Memory}

Planning on STAG is consistently faster than planning directly on the grid. As seen in Table \ref{tab:stag_results}, on the synthetic cave environment, median planning time decreases from 658 ms on the grid to 82.5 ms on STAG, corresponding to an \(8\times\) speed-up. On the long synthetic mine map, planning time decreases from 1797.2 ms to 181.2 ms, corresponding to an \(9.9\times\) speed-up. The DARPA-style maps show the same trend, with speed-ups of \(3.8\times\) and \(7.6\times\).

Peak query memory is also reduced in all benchmark categories. As shown in Table~\ref{tab:stag_results}, the memory-reduction factor ranges from \(2.1\times\) for Synthetic mines to \(15.4\times\) for the DARPA \(0.25\,\mathrm{m}\) environment. The Long synthetic mine achieves a \(6.9\times\) reduction, decreasing from \(4158.3\,\mathrm{KiB}\) to
\(603.4\,\mathrm{KiB}\).

These results support the second claim of the paper: for the same A* planner and the same start-goal queries, STAG reduces both planning time and peak query memory compared with dense grid planning.

\subsection{Path Quality}

The geometric length of the STAG paths remains close to the grid baseline. For Synthetic caves, Synthetic mines, and the Long synthetic
mine, the median relative path-length differences are \(+7.6\%\), \(+0.9\%\), and \(-2.9\%\), respectively. For the two DARPA-derived
maps, the corresponding increases are \(+6.0\%\) and \(+6.2\%\). The negative value for the Long synthetic mine does not necessarily
indicate a better solution, because the grid and graph representations use different motion primitives.

Traversability penalty shows a clearer abstraction trade-off. The median relative differences are \(+6.9\%\), \(+3.4\%\), and \(+2.0\%\)
for Synthetic caves, Synthetic mines, and the Long synthetic mine, respectively. The increases are substantially larger on the DARPA-derived maps, reaching \(+50.9\%\) at \(0.5\,\mathrm{m}\) resolution and \(+38.2\%\) at \(0.25\,\mathrm{m}\) resolution.

\subsection{Component Ablation}

A separate component ablation is conducted on four representative single-map benchmarks comprising 1,700 grid-valid start-goal queries: 500 queries on the representative synthetic mine, 200 on the Long synthetic mine, and 500 on each DARPA-derived map. The variants include the topological backbone alone, topology with region nodes, topology with transition nodes, and the full STAG representation. The same grid-valid start-goal pairs are reused for every variant. A failure denotes a query for which the grid baseline finds a valid path but the graph variant does not.

\begin{table}[t]
\centering
\caption{Component ablation on four representative maps comprising 1,700 grid-valid start-goal queries.}
\label{tab:ablation}
\resizebox{\columnwidth}{!}{
\begin{tabular}{lrrl}
\hline
\textbf{Variant} &
\textbf{Success} &
\textbf{Failures} &
\textbf{Main observation} \\
\hline
Topology only &
1560/1700 &
140 &
Sparse but less robust \\
Topology + region &
1699/1700 &
1 &
Restores connectivity \\
Topology + transition &
1605/1700 &
95 &
Adds detail, limited coverage \\
Full STAG &
1699/1700 &
1 &
Robust coverage and transition \\
\hline
\end{tabular}
}
\end{table}

Table~\ref{tab:ablation} shows that the topological backbone alone is not sufficient for robust planning over all valid grid queries. Although topology-only graphs are highly compact, they fail when the skeleton does not provide enough accessible structure for arbitrary start-goal attachment or terrain-aware routing. Adding region nodes removes nearly all failures, showing that homogeneous traversability representatives are essential for robust endpoint coverage and graph-query success. Region nodes are the primary contributor to query coverage and traversability-aware path quality. Transition nodes encode cost discontinuities and provide additional structure for downstream planners; however, the current A* weighting does not consistently exploit this structure to reduce accumulated traversability penalty.

% The full STAG representation therefore benefits primarily from combining the topological backbone with region-level traversability structure.

\subsection{Discussion}

The results support the efficiency claims of the paper and clarify the representation trade-off. STAG consistently reduces the planning state space, planning time, and peak query memory, while keeping path length close to the grid baseline. Traversability cost, however, depends on the structure of the underlying risk field: it is nearly preserved in the synthetic mine maps but increases on the DARPA-style maps.

This result clarifies the role of STAG. The proposed representation is not designed to exactly reproduce the optimal path of the dense grid. Instead, it provides a sparse global abstraction that enables faster and lighter planning. STAG is therefore most suitable for global planning, long-range navigation, or repeated path queries over large maps. For applications requiring precise local terrain-cost optimization, STAG should be combined with a local planner operating on the original costmap around the robot or around the selected global path.

% No comparison with TRG-Planner or other dedicated traversability graph planners is included in the current evaluation. Consequently, the present results should not be interpreted as demonstrating superiority over these methods. The benchmark supports the more specific claim that, for the same A* planner and the same start-goal queries, the proposed STAG representation is more computationally efficient than planning directly on the grid while maintaining a measurable path-quality trade-off.

The evaluation isolates the effect of the environment representation by using the same A* algorithm, cost function, and planning queries for the grid and STAG. Dedicated traversability planners combine their own representation, sampling, connection, and search strategies, so a direct comparison would mix several contributions. The claims of this work are
therefore limited to showing that STAG provides a sparser planning space with lower query time and memory than the original grid, while introducing a measurable path-quality trade-off.

\subsection{Limitations and Scope}

The proposed representation is intended for sparse global planning rather than dense local terrain optimization. Since STAG compresses the input costmap into topological, regional, and transition-level graph elements, fine local variations in traversability may be lost. This effect is visible in the DARPA-style maps, where path length remains close to the grid baseline but accumulated traversability cost increases. In practice, STAG should therefore be combined with a local planner or trajectory optimizer that operates on the original costmap near the robot or around the selected global route.

The current evaluation isolates the effect of the representation by comparing grid-based and STAG-based A* on the same start-goal queries. It does not claim superiority over dedicated traversability graph planners, nor does it evaluate dynamic replanning or online graph updates. Future extensions should investigate adaptive refinement in high-risk regions, incremental graph maintenance as new map data arrives, and hybrid graph-grid planning strategies.

\section{Conclusion}

This paper introduced STAG, a Sparse Traversability-Aware Graph representation for converting grid-based traversability costmaps into compact graphs for global path planning. STAG combines a medial-axis topological backbone with traversability-aware region and transition nodes, producing a sparse graph that stores geometric, clearance, and terrain-cost attributes on its edges.

The experimental evaluation compared A* planning on the original traversability grid with A* planning on the proposed STAG representation. Across synthetic cave and mine environments and two DARPA-derived maps, STAG consistently reduced query time and peak Python allocations, yielding approximately \(3\times\) to \(10\times\) faster planning and \(2\times\) to \(15\times\) lower peak allocations than grid-based A*. These gains are obtained while keeping geometric path
length close to the grid baseline.

The results also show the expected trade-off introduced by sparse abstraction: accumulated traversability penalty increases on the more complex DARPA-derived maps. The component ablation shows that region nodes are the main contributor to endpoint coverage, graph-query success, and traversability-aware path-quality recovery, whereas transition nodes primarily encode additional terrain-cost structure that is not consistently exploited by the current A* weighting.

STAG should therefore be interpreted as a lightweight global planning representation rather than a complete replacement for dense local costmap planning. Future work will focus on adaptive graph refinement in areas where traversability information is poorly approximated, hybrid graph-grid planning, and direct comparison with dedicated traversability graph planners.
The project is open source and you can find the code at this \href{https://anonymous.4open.science/r/STAG-C802/README.md}{link}.

% \newpage

\addtolength{\textheight}{-12cm}   % This command serves to balance the column lengths

%%%%%%%%%%%%%%%%%%%%%%%%%%%%%%%%%%%%%%%%%%%%%%%%%%%%%%%%%%%%%%%%%%%%%%%%%%%%%%%%

%%%%%%%%%%%%%%%%%%%%%%%%%%%%%%%%%%%%%%%%%%%%%%%%%%%%%%%%%%%%%%%%%%%%%%%%%%%%%%%%

%%%%%%%%%%%%%%%%%%%%%%%%%%%%%%%%%%%%%%%%%%%%%%%%%%%%%%%%%%%%%%%%%%%%%%%%%%%%%%%%

%%%%%%%%%%%%%%%%%%%%%%%%%%%%%%%%%%%%%%%%%%%%%%%%%%%%%%%%%%%%%%%%%%%%%%%%%%%%%%%%

\end{document}